\documentclass{article}
\usepackage{ijcai24}

\usepackage{times}
\usepackage{soul}
\usepackage{url}
\usepackage[hidelinks]{hyperref}
\usepackage[utf8]{inputenc}
\usepackage[small]{caption}
\usepackage{graphicx}
\usepackage{amsmath}
\usepackage{amsthm}
\usepackage{booktabs}
\usepackage{algorithm}
\usepackage{algorithmic}
\usepackage[switch]{lineno}
\usepackage{subfigure}

\usepackage{natbib}

\newtheorem{theorem}{Theorem}
\newtheorem{lemma}{Lemma}

\newtheorem{remark}{Remark}
\newtheorem{definition}{Definition}
\newtheorem{assumption}{Assumption}
\usepackage{subfigure}
\usepackage{multicol}
\usepackage{amsfonts}

\title{Fine-grained Analysis of Stability and Generalization for Stochastic Bilevel Optimization}

\author{
Xuelin Zhang$^1$
\and 
Hong Chen$^{1,2,}$\footnote{Corresponding author.}
\and
Bin Gu$^3$
\and
Tieliang Gong$^4$
\And
Feng Zheng$^5$\\
\affiliations
$^1$College of Informatics, Huazhong Agricultural University, Wuhan 430070, China\\
$^2$Engineering Research Center of Intelligent Technology for Agriculture, Ministry of Education, Wuhan 430070, China\\
$^3$School of Artificial Intelligence, Jilin University, Jilin 130012, China\\
$^4$School of Computer Science and Technology, Xi’an Jiaotong University, Xi’an 710049, China\\
$^5$Department of Computer Science and Engineering, Southern University of Science and Technology, Shenzhen 518055, China\\
\emails
zhangxuelin@webmail.hzau.edu.cn,
chenh@mail.hzau.edu.cn}

\begin{document}

\maketitle

\begin{abstract}
Stochastic bilevel optimization (SBO) has been integrated into many machine learning paradigms recently, including hyperparameter optimization, meta learning, and reinforcement learning. Along with the wide range of applications, there have been numerous studies on the computational behavior of SBO. However, the generalization guarantees of SBO methods are far less understood from the lens of statistical learning theory. 
In this paper, we provide a systematic generalization analysis of the first-order gradient-based bilevel optimization methods. Firstly, we establish the quantitative connections between the on-average argument stability and the generalization gap of SBO methods. Then, we derive the upper bounds of on-average argument stability for single-timescale stochastic gradient descent (SGD) and two-timescale SGD, where three settings (nonconvex-nonconvex (NC-NC), convex-convex (C-C), and strongly-convex-strongly-convex (SC-SC)) are considered respectively. Experimental analysis validates our theoretical findings. 
Compared with the previous algorithmic stability analysis, our results do not require reinitializing the inner-level parameters at each iteration and are applicable to more general objective functions.
\end{abstract}

\section{Introduction}
In this paper, we focus on establishing stability and generalization analysis for the stochastic bilevel optimization (SBO)  \citep{BrackenM73a,hyper_convergence1, DBLP:conf/nips/BaoWLZZ21} defined as follows:
\begin{equation}\label{bilevel_problem}
\begin{aligned}
\min_{x \in \mathbb{R}^{d_1}} R(x) &=F\left(x, y^*(x)\right)
:=\mathbb{E}_{\xi}\left[f\left(x, y^*(x) ; \xi\right)\right] \\
\text { s.t. } y^*(x) &=\arg \min _{y \in \mathbb{R}^{d_2}}\left\{G(x, y):=\mathbb{E}_\zeta[g(x, y; \zeta)]\right\},
\end{aligned}
\end{equation}
where $d_1, d_2 \in \mathbb{N}^+$, the outer objective function $f$ and the inner objective function $g$ are both continuous and differentiable,  $\xi,\zeta$ are samples drawn from the validation set and training set, respectively.

For this bilevel optimization scheme, we often call $\min _{y \in \mathbb{R}^{d_2}} \mathbb{E}_\zeta[g(x, y; \zeta)]$ as the inner (or lower-level) problem, and name $\min_{x \in \mathbb{R}^{d_1}} \mathbb{E}_{\xi}\left[f\left(x, y^*(x); \xi\right)\right]$ as the outer (or upper-level) problem. The goal of \eqref{bilevel_problem} is
to minimize the outer objective function $R(x)$ (also $F(x,y^*(x))$) with respect to (w.r.t.) the model parameter $x$, where parameter  $y^*(x)$ is derived from the inner minimization formulation.

The SBO formulation in \eqref{bilevel_problem}, stemming from \citep{BrackenM73a}, has attracted increasing attention in many machine learning applications, including hyper-parameter optimization \citep{DBLP:conf/icml/FranceschiDFP17,franceschi2018bilevel,DBLP:journals/corr/LorraineD18,DBLP:conf/iclr/MacKayVLDG19,DBLP:journals/jmlr/OkunoTKW21,zhang2023robust}, generative adversarial learning \citep{DBLP:journals/corr/PfauV16}, 
meta learning \citep{franceschi2018bilevel,meta1,DBLP:conf/iclr/ZugnerG19,meta2,meta3}, and reinforcement learning \citep{DBLP:conf/nips/TschiatschekGHD19}.
Indeed, there are numerous computational methods for implementing this bilevel optimization scheme, as well as theoretical works on convergence analysis of optimization \citep{DBLP:journals/corr/LiGH20,hyper_convergence1}. 
However, the generalization analysis of SBO is still far less understood from the viewpoint of statistical learning theory (STL), e.g., algorithmic stability and generalization analysis \citep{DBLP:conf/icml/HardtRS16,DBLP:conf/icml/LeiY20,DBLP:conf/nips/BaoWLZZ21}.

Stability-based generalization analysis can be traced back to the 1970s \citep{Ann.Stat/RogersW1978} and has achieved rapid developments in STL, see e.g.,  \citep{DBLP:journals/jmlr/BousquetE02,DBLP:journals/jmlr/ElisseeffEP05,DBLP:conf/icml/HardtRS16,liu2017algorithmic,DBLP:conf/icml/LeiY20,DBLP:conf/nips/LeiLY21,deng2021toward,kuzborskij2018data}.
To match the characterizations of various algorithms, different definitions of algorithmic stability have been formulated (including the uniform stability \citep{DBLP:journals/jmlr/BousquetE02}, uniform argument stability \citep{liu2017algorithmic}, locally elastic stability \citep{deng2021toward}, on-average stability \citep{kuzborskij2018data} and on-average argument stability \citep{DBLP:conf/icml/LeiY20}) to better investigate their generalization bounds. The on-average argument stability was proposed in \citep{DBLP:conf/icml/LeiY20} to establish the fine-grained generalization analysis of single-level pointwise stochastic gradient descent (SGD).
Subsequently, Lei et al. extended the stability-based generalization assessment to pairwise SGD \citep{DBLP:conf/nips/LeiLY21}, providing systematic strategies to better balance generalization error and optimization error.
As far as we know, there is only one study exploring the generalization analysis of SBO \citep{DBLP:conf/nips/BaoWLZZ21}, which presents an expectation generalization bound w.r.t. the validation set via
the uniform stability approach. 
However, the theoretical analysis of \citep{DBLP:conf/nips/BaoWLZZ21} is limited to unrolled differentiation (UD) based algorithms with re-initialization in inner-level for hyper-parameter optimization, which may not be applicable to other commonly used optimization algorithms, e.g., single timescale SGD (SSGD) \citep{zhou2022probabilistic,liu2022bome,chen2022single} and two timescale SGD (TSGD) \citep{zhou2022probabilistic,liu2022bome,hong2023two}.
Therefore, it is important to further investigate the generalization guarantees of the general SBO formulation to cover a wider range of bilevel optimization algorithms. 

To address the aforementioned issue, this paper establishes the fine-grained stability and generalization analysis for general first-order bilevel optimization methods.
Our main contributions are summarized as follows:
\begin{itemize}
\item Firstly, we establish the quantitative connection between the generalization gap of bilevel optimization methods and on-average argument stability. Especially for the $l_2$ on-average argument stability, the derived stability-based generalization bounds involve the empirical risks, which are consistent with the previous analysis for single-level optimization \citep{DBLP:conf/icml/LeiY20,DBLP:conf/nips/LeiLY21}.

\item Secondly, this paper provides several stability bounds for bilevel optimization methods associated with both SSGD and TSGD algorithms, under different objective function conditions (i.e., SC-SC, C-C, NC-NC). Moreover, we extend the results to a more general setting by relaxing the restrictions on the optimization objective (e.g., Lipschitz continuity and smoothness assumptions). 
As far as we know, this is the first systemic generalization analysis for first-order SGD-based bilevel optimization under the low-noise setting.

\item Finally, we conduct experimental evaluations for bilevel optimization methods, including hyperparameter optimization. Empirical results validate our theoretical findings about the relationship between the generalization gap and the size of the validation set, as well as the maximum value of inner (outer) iterations. 
\end{itemize}

To better evaluate our results, we  compare them with the most related work on stability and generalization analysis \citep{DBLP:conf/nips/BaoWLZZ21} from the following perspectives:
\begin{itemize}
    \item
\emph{Optimization strategy.} The previous UD-based hyperparameter optimization (Algorithm 1 in \citep{DBLP:conf/nips/BaoWLZZ21}) requires reinitialization in the inner-level parameters before each iteration. Different from this special case, this paper considers SBO algorithms in which the parameters at the inner and outer levels are both updated continuously (e.g., SSGD \citep{zhou2022probabilistic,chen2022single} and TSGD \citep{zhou2022probabilistic,liu2022bome,hong2023two}). The iteration strategy matching our analysis has been used extensively in practice \citep{hyper_convergence1,liu2022bome,ghadimi2018approximation}. Especially for the theoretical analysis of TSGD, it is challenging to handle gradient summation during inner iterations, and the previous analysis technique \citep{DBLP:conf/nips/BaoWLZZ21} cannot be extended to this case directly.

\item\emph{Analysis tool.} Different from uniform stability used in  \citep{DBLP:conf/nips/BaoWLZZ21}, this paper develops the analysis technique of on-average argument stability to provide the fine-grained generalization bounds under low noise settings, where the stability bounds involve a weighted sum of empirical risks instead of the uniform Lipschitz constants.

\item\emph{Conditions of objective functions.} 
Similar to the previous stability analysis in \citep{lei2020sharper,shen2020stability,zhou2022understanding}, the objective functions in \citep{DBLP:conf/nips/BaoWLZZ21} are assumed to be bounded, third-order continuously differentiable, and smooth. Here, we merely need the bilevel objective functions to be nonnegative, smooth and Lipschitz continuous, where the last condition for the outer-level function can be further removed by the $l_2$ on-average argument stability. Detailed stability results have been derived for both SSGD and TSGD algorithms under NC-NC, C-C and SC-SC settings. In addition, we also establish generalization bounds by replacing the smooth condition with the weaker H$\ddot{o}$lder continuous assumption.  
\end{itemize}

\section{Problem Formulation} \label{section2}

Given distributions $\mathbb{D}_1$, $\mathbb{D}_2$, we get the  validation set \begin{equation*} D_{m_1} := \{\xi_i\}_{i=1}^{m_1} \sim \mathbb{D}_1^{m_1}
\end{equation*}  and
the training set 
\begin{equation*} D_{m_2} := \{\zeta_i\}_{i=1}^{m_2} \sim \mathbb{D}_2^{m_2}\end{equation*}  
by independent sampling, where $m_1$ and $m_2$ are the sample sizes. This paper focuses on the outer-level population risk w.r.t $\mathbb{D}_1$ and empirical risk w.r.t $D_{m_1}$ \footnote{Thus we consider adding corruptions to $D_{m_1}$ to access the generalization behavior of the meta-learner \citep{thrun1998lifelong} at upper level.}, which are defined respectively as 
\begin{equation*} 
\begin{aligned}
& R\left(x, y\right)=\mathbb{E}_{\xi \sim \mathbb{D}_1}[f(x, y(x) ; \xi)]\\
\mbox{and} ~~ & R_{D_{m_1}}\left(x, y\right)=\frac{1}{m_1} \sum_{i=1}^{m_1}\left[f\left(x, y(x) ; \xi_i\right)\right],
\end{aligned}
\end{equation*}
where $f: \mathbb{R}^{d_1} \times \mathbb{R}^{d_2} \rightarrow \mathbb{R}$ is an objective function and $y(x)$ is the inner model parameter given the outer model parameter $x$ (also see \eqref{bilevel_problem}).

Let $(x, y(x))$ in \eqref{bilevel_problem} be estimated by a stochastic algorithm $A$ with data $D_{m_1}, D_{m_2}$, i.e. $A\left(D_{m_1}, D_{m_2}\right)$.  
Similar to the previous works \citep{DBLP:conf/nips/BaoWLZZ21,hoffer2017train,keskar2017on}, in order to evaluate the approximated searching of hyperparameters, we define
\begin{equation}\label{generalization_gap}
\mathbb{E}_{A, D_{m_1}, D_{m_2}}\left[R\left(A\left(D_{m_1}, D_{m_2}\right)\right)-R_{D_{m_1}}\left(A\left(D_{m_1}, D_{m_2}\right)\right)\right]
\end{equation}
in the upper (outer) level as the generalization gap of $A$, which measures the difference between the population risk $R(A)$ and the empirical risk $R_{D_{m_1}}(A)$.

The following conditions have been used to characterize the theoretical properties of objective functions in \eqref{bilevel_problem}.

\begin{definition}\label{definition_1}
(Joint Lipschitz Continuity \citep{hyper_convergence1,liu2022bome}). An objective function $f$ is jointly $L_f$-Lipschitz over $\mathbb{R}^{d_1} \times \mathbb{R}^{d_2}$,  if there holds
\begin{equation*}
\left| f(x,y;\xi)-f( x^{\prime},y^{\prime}; \xi)\right| \leq L_f \sqrt{\left\|x-x^{\prime}\right\|_2^2+\left\|y-y^{\prime}\right\|_2^2}
\end{equation*}
for any  $(x,y), (x^{\prime},y^{\prime})\in \mathbb{R}^{d_1} \times \mathbb{R}^{d_2} ,\xi\sim \mathbb{D}_1$.
\end{definition}

\begin{definition}\label{definition_2}
(Joint Smoothness \citep{DBLP:conf/icml/LeiYYY21}). An objective function $f$ is $\ell_f$-smooth over $\mathbb{R}^{d_1} \times \mathbb{R}^{d_2}$,  if
\begin{equation*}
\|\nabla f(x,y;\xi)-\nabla f( x^{\prime},y^{\prime};\xi)\|_2 \leq \ell_f \sqrt{\left\|x-x^{\prime}\right\|_2^2+\left\|y-y^{\prime}\right\|_2^2}
\end{equation*}
for any  $(x,y), (x^{\prime},y^{\prime})\in \mathbb{R}^{d_1} \times \mathbb{R}^{d_2} ,\xi\sim \mathbb{D}_1$,
\end{definition}

\begin{definition}\label{definition_3}
(Strong Convexity). A function $\psi$ is $\mu$-strongly-convex over a set $X$, if $\forall t, t^{\prime} \in X$,
\begin{equation*}
\psi(t^\prime)+\langle\nabla \psi(t^\prime), t-t^\prime\rangle+\frac{\mu}{2}\|t-t^\prime\|_2^2 \leq \psi(t).
\end{equation*}
\end{definition}

\begin{definition}\label{definition_4}
(H$\ddot{o}$lder Continuity). Let $\tau>0, \alpha \in[0,1]$. Gradient $\nabla f$ is $(\alpha,\tau)$-H$\ddot{o}$lder continuous over $\mathbb{R}^{d_1} \times \mathbb{R}^{d_2}$, if there holds
\begin{equation*}
\|\nabla f(x,y;\xi)-\nabla f( x^{\prime},y^{\prime}; \xi)\|_2 \leq \tau \left\|\begin{array}{l}
x-x^{\prime} \\
y-y^{\prime}
\end{array}\right\|_2^\alpha
\end{equation*}
for all $(x,y), (x^{\prime},y^{\prime})\in \mathbb{R}^{d_1} \times \mathbb{R}^{d_2}$ and  $\xi\sim \mathbb{D}_1$.
\end{definition}

The above conditions for objective functions have been used extensively in convergence analysis for bilevel optimization \citep{hyper_convergence1,ghadimi2018approximation,liu2022bome} and stability-based generalization analysis for single-level optimization methods \citep{DBLP:conf/icml/HardtRS16,DBLP:conf/icml/LeiYYY21}.
Moreover,  the H$\ddot{o}$lder continuity is much weaker than the Lipschitz continuity and smoothness \citep{DBLP:conf/icml/LeiY20,nesterov2015universal}. If Definition \ref{definition_4} holds with $\alpha=1$, then $f$ is smooth (see Definition \ref{definition_2}). And if Definition \ref{definition_4} holds with $\alpha=0$, $f$ becomes Lipschitz continuous as in Definition \ref{definition_1} and can be non-differentiable \citep{DBLP:conf/icml/LeiY20}. The objective functions satisfying Definition \ref{definition_4} include the mean absolute function, the hinge function and some of their variants \citep{DBLP:conf/icml/LeiY20,steinwart2008support}.

\begin{definition}\label{definition_5}
(On-average Argument Stability \citep{DBLP:conf/icml/LeiY20}). Let $D_{m_1}=\left\{z_1, \ldots, z_{m_1}\right\}$ and $\widetilde{D}_{m_1}=\left\{\tilde{z}_1, \ldots, \tilde{z}_{m_1}\right\}$ be two sets drawn independently from distribution $ \mathbb{D}_1^{m_1}$. For any $i=1, \ldots, m_1$, define $D^{(i)}=$ $\left\{z_1, \ldots, z_{i-1}, \tilde{z}_i, z_{i+1}, \ldots, z_{m_1}\right\}$. Denote the $\mathbb{E}$ as the expectation of $\mathbb{E}_{D_{m_1}, D_{m_2},\widetilde{D}_{m_1}, A}$.
We say a randomized algorithm $A$ is $l_1(\beta)$ on-average argument stable if
\begin{equation*}
\mathbb{E}\left[\frac{1}{m_1} \sum_{i=1}^{m_1}\left\|A(D_{m_1},D_{m_2})-A\left(D_{m_1}^{(i)},D_{m_2}\right)\right\|_2\right] \leq \beta,
\end{equation*}
and $l_2(\beta^2)$ on-average argument stable if
\begin{equation*} 
\begin{aligned}
\mathbb{E}\left[\frac{1}{m_1} \sum_{i=1}^{m_1}\left\|A(D_{m_1},D_{m_2})-A\left(D_{m_1}^{(i)},D_{m_2}\right)\right\|_2^2\right] \leq \beta^2.
\end{aligned}
\end{equation*}
\end{definition}
\begin{remark}
The on-average argument stability measures the average sensitivity (stability) of the learning algorithm's output parameters when at most one validation sample is changed. Definition \ref{definition_5} is different from Definition 1 in  \citep{DBLP:conf/nips/BaoWLZZ21},  where the uniform stability is evaluated by the drift of prediction error of the hyperparameter optimization algorithm, and the boundedness of the loss function is often required. 
\end{remark}

Based on the above definitions, we introduce the requirements of $f,g$ in our analysis.

\begin{assumption}\label{assumption_1}
(Outer Function Assumption). Assume that the outer objective function $f$ in \eqref{bilevel_problem} satisfies

(\textrm{I}) $f$ is  jointly $L_f$-Lipschitz.

(\textrm{II}) $f$ is nonnegative, continuously differentiable and $\ell_f$-smooth.
\end{assumption}
\begin{assumption}\label{assumption_2}
(Inner Function Assumption). Assume that the inner objective function $g$ in \eqref{bilevel_problem} satisfies 

(\textrm{I}) $g$ is jointly $L_g$-Lipschitz.

(\textrm{II}) $g$ is continuously differentiable and $\ell_g$-smooth.
\end{assumption}

\section{Quantitative Relationship between Generalization and Stability} \label{section3}

This section states that the generalization gap of \eqref{bilevel_problem} can be bounded by the on-average argument stability. 
Before providing the detailed conclusion of Theorem 1, we first introduce the self-bounding property definition.

\begin{lemma} \label{self-bounding}
(Self-bounding property). Assume that for all $z \in \mathcal{D}$, the map ${w} \mapsto f({w} ; z)$ is nonnegative, and ${w} \mapsto \partial f({w} ; z)$ is $(\alpha, \tau)$-H$\ddot{o}$lder continuous with $\alpha \in[0,1]$. 
Then we have
\begin{equation*}
\|\partial f({w}, z)\|_2 \leq c_{\alpha, \tau} f^{\frac{\alpha}{1+\alpha}}({w}, z), \quad \forall {w} \in \mathbb{R}^d, z \in \mathcal{D},
\end{equation*}
where $c_{\alpha, \tau}= \begin{cases}(1+1 / \alpha)^{\frac{\alpha}{1+\alpha}} \tau^{\frac{1}{1+\alpha}}, & \text { if } \alpha>0 \\ \sup _z\|\partial f(0 ; z)\|_2+\tau, & \text { if } \alpha=0\end{cases}$.
\end{lemma}

The self-bounding property of $f$ with ($\alpha,\tau$)-H$\ddot{o}$lder continuous (sub)gradient contains the specific Lipschitz continuous ($\alpha = 0$) and smoothness ($\alpha = 1$) conditions \citep{DBLP:conf/icml/LeiYYY21}.

\begin{theorem}\label{theorem1}
(\textrm{I}) If algorithm $A$ is $l_1(\beta)$ on-average argument stable in expectation and the outer-level function $f$ is $L_f$-Lipschitz continuous w.r.t. $(x,y) \in \mathbb{R}^{d_1} \times \mathbb{R}^{d_2}$, denote $\mathbb{E}$ as $\mathbb{E}_{A, D_{m_1}, D_{m_2}}$, there holds 
\begin{equation*}
\begin{aligned}
&|\mathbb{E}\left[R\left(A\left(D_{m_1}, D_{m_2}\right)\right)-R_{D_{m_1}}\left(A\left(D_{m_1}, D_{m_2}\right)\right)\right]| 
\leq L_f \beta.
\end{aligned}
\end{equation*}

(\textrm{II}) If algorithm $A$ is $l_2(\beta^2)$ on-average argument stable in expectation and f is nonnegative and $\ell_f$-smooth w.r.t. $(x,y) \in \mathbb{R}^{d_1} \times \mathbb{R}^{d_2}$, denote $\mathbb{E}$ as $\mathbb{E}_{A, D_{m_1}, D_{m_2}}$, then 
\begin{equation*}\small
\centering
\begin{aligned}\label{l2_stability}
& \mathbb{E}\left[R\left(A\left(D_{m_1}, D_{m_2}\right)\right)-R_{D_{m_1}}\left(A\left(D_{m_1}, D_{m_2}\right)\right)\right] \\
\leq  &  \frac{\ell_f}{\gamma} \mathbb{E}\left[R_{D_{m_1}}(A(D_{m_1},D_{m_2}))\right] 
+\frac{(\ell_f+\gamma) \beta^2}{2} ,
\end{aligned}
\end{equation*}
where the constant $\gamma>0$. 

(\textrm{III}) If algorithm $A$ is $l_2(\beta^2)$ on-average argument stable in expectation, f is nonnegative and $(\alpha,\tau)$-H$\ddot{o}$lder continuous w.r.t. $(x,y) \in \mathbb{R}^{d_1} \times \mathbb{R}^{d_2}$ with $\alpha \in [0,1]$, then 
\begin{eqnarray*}
&& \mathbb{E}\left[R\left(A\left(D_{m_1}, D_{m_2}\right)\right)-R_{D_{m_1}}\left(A\left(D_{m_1}, D_{m_2}\right)\right)\right] \\
&\leq&  \frac{c_{\alpha, \tau}^2}{2 \gamma} \mathbb{E}\left[R^{\frac{2 \alpha}{1+\alpha}}(A\left(D_{m_1}, D_{m_2}\right))\right] + \frac{\gamma}{2} \beta^2
\end{eqnarray*}
for $D_{m_1} \sim \mathbb{D}_1^{m_1}$ and $D_{m_2} \sim \mathbb{D}_2^{m_2}$,
where the constant $\gamma>0$.
\end{theorem}

\begin{remark}
Theorem \ref{theorem1} validates the connection between on-average argument stability and the generalization gap. Especially, the smoothness assumption is further relaxed by the H$\ddot{o}$lder continuity in Theorem \ref{theorem1}(III).
\end{remark}

\begin{remark}
Different from the uniform stability technique employed in \citep{DBLP:conf/nips/BaoWLZZ21}, the on-average argument stability further exploits the Lipschitz continuous ($L_f$) or smooth properties ($\ell_f$) of the objective function as well as the stability parameter ($\beta$) to bound the algorithmic generalization gap. Especially, there is a trade-off between the empirical risk and the algorithmic stability bound.
\end{remark}

\begin{algorithm}[!t]
\caption{Computing algorithm of SSGD}
\textbf{Input}: Validation data $D_{m_1} = \{\xi_i\}_{i=1}^{m_1}$ and training set $D_{m_2} = \{\zeta_i\}_{i=1}^{m_2}$, the total number of iterations $K$, step sizes $\eta_x$, $\eta_y$. \\
\textbf{Initialization}: $x_0$ and $y_0$.

\begin{algorithmic}[1] 
\FOR{$k=1$ to $K-1$}{
\STATE Uniformly sample $\xi_i \in D_{m_1}$ and $\zeta_i \in D_{m_2}$:\\
\STATE $y_{k+1}=y_k-\eta_y \nabla_{y} g\left(x_k, y_k\left(x_k\right) ; \zeta_i\right)$\\
\STATE $x_{k+1}=x_k-\eta_x \nabla_x f\left(x_k, y_k\left(x_k\right) ; \xi_i\right)$\\
}
\ENDFOR
\end{algorithmic}
\textbf{Output}: $x_K$ and $y_K$.
\label{algorithm-SSGD}
\end{algorithm}

\begin{algorithm}[t]
\caption{Computing algorithm of TSGD}
\textbf{Input}: Validation data $D_{m_1} = \{\xi_i\}_{i=1}^{m_1}$ and training set $D_{m_2} = \{\zeta_i\}_{i=1}^{m_2}$, the total number of inner iterations $T$ and outer iterations $K$, step sizes $\eta_x$ and $\eta_y$.\\
\textbf{Initialization}: $x_0$ and $y_0^0$.

\begin{algorithmic}[1] 
\FOR{$k=0$ to $K-1$}{
    \FOR{$t=0$ to $T-1$}{
    \STATE Uniformly sample $\zeta_i \in D_{m_2}$:\\
    \STATE $y_{k}^{t+1}=y_k^t-\eta_y \nabla_{y} g\left(x_k, y_k^t\left(x_k\right) ; \zeta_i\right)$\\
    }\ENDFOR
    \STATE Uniformly sample $\xi_i \in D_{m_1}$:\\
    \STATE $x_{k+1}=x_k-\eta_x \nabla_x f\left(x_k, y_k^T\left(x_k\right) ;\xi_i\right)$\\
    \STATE $y_{k+1}^0 = y_k^T$\\
}
\ENDFOR
\end{algorithmic}
\textbf{Output}: $x_K$ and $y_K^0$.
\label{algorithm-TSGD}
\end{algorithm}

\begin{algorithm}[!t]
\caption{Computing algorithm of UD \citep{DBLP:conf/nips/BaoWLZZ21}}
\textbf{Input}: Validation data $D_{m_1} = \{\xi_i\}_{i=1}^{m_1}$ and training set $D_{m_2} = \{\zeta_i\}_{i=1}^{m_2}$, the total number of inner iterations $T$ and outer iterations $K$, step sizes $\eta_x$ and $\eta_y$.\\
\textbf{Initialization}: $x_0$ and $y^0$.

\begin{algorithmic}[0] 
\FOR{$k=0$ to $K-1$}{
    \STATE $ y_k^0 = y^0$\\
    \FOR{$t=0$ to $T-1$}{
    \STATE Uniform sampling $\zeta_i \in D_{m_2}$:\\
    \STATE $y_{k}^{t+1}=y_k^t-\eta_y \nabla_{y} g\left(x_k, y_k^t\left(x_k\right) ; \zeta_i\right)$\\
    }\ENDFOR
    \STATE Uniform sampling $\xi_i \in D_{m_1}$:\\
    \STATE $x_{k+1}=x_k-\eta_x \nabla_x f\left(x_k, y_k^T\left(x_k\right) ;\xi_i\right)$\\
    \STATE $y_{k+1}^0 = y_k^T$\\
}
\ENDFOR
\end{algorithmic}
\textbf{Output}: $x_K$ and $y_K^0$.
\label{algorithm-UD}
\end{algorithm}

\begin{remark}
There are several advantages of $l_2$ on-average argument stability in Theorem \ref{theorem1} (II), where Assumption \ref{assumption_1}(\textrm{I}) is removed and the low noise assumption can be used to obtain a fine-grained result instead of the Lipschitz constant \citep{DBLP:conf/icml/LeiY20}. If algorithm $A$ is $l_2(\beta^2)$ on-average argument stable, then we derive the upper bound of generalization gap with $\sqrt{2 \ell_f \mathbb{E}\left[R_{D_{m_1}}(A(D_{m_1},D_{m_2}))\right]} \beta + \ell_f \beta^2 / 2$ by taking $\gamma=\sqrt{2 \ell_f \mathbb{E}\left[R_{D_{m_1}}(A(D_{m_1},D_{m_2}))\right]} / \beta$.
Moreover, if the output model achieves a small empirical risk (e.g., low noise assumption $\mathbb{E}\left[R_{D_{m_1}}(A(D_{m_1},D_{m_2}))\right] = \mathcal{O}(m_1^{-1})$), we get that $\mathbb{E}\left[R\left(A\left(D_{m_1}, D_{m_2}\right)\right)-R_{D_{m_1}}\left(A\left(D_{m_1}, D_{m_2}\right)\right)\right] = \mathcal{O}\left(\beta^2+\beta / \sqrt{m_1}\right)$.
\end{remark}

\section{Stability Analysis for Stochastic Bilevel Optimization}\label{section4}
To solve the bilevel optimization formulation \eqref{bilevel_problem}, some gradient-based algorithms are designed based on the single timescale or two timescale strategies \citep{hyper_convergence1,chen2022single,liu2020generic,liu2022bome,zhou2022probabilistic}.
In the following, we introduce the computational approaches for \eqref{bilevel_problem} (SSGD in Algorithm \ref{algorithm-SSGD} and TSGD in Algorithm \ref{algorithm-TSGD}) and then assess their generalization by presenting their algorithmic stability bounds. 

\subsection{Stability and Generalization Analysis for SSGD}
Let $\eta_x$ and $\eta_y$ be the step sizes for updating $x$ and $y$. According to Theorem \ref{theorem1}, the on-average argument stable metrics in Definition \ref{definition_5} for SSGD algorithm $A$ with $K$ iterations $\left\|A(D_{m_1},D_{m_2})-A\left(D_{m_1}^{(i)},D_{m_2}\right)\right\|_2$ can be measured by  
\begin{equation*}
\sqrt{\|x_{K}-x_{K}^{(i)}\|_2^2 +  \|y_{K}-y_{K}^{(i)}\|_2^2}.
\end{equation*}

\begin{table*}[!t]
\centering 
\caption{Summary of the generalization bounds under different settings. For briefly,  $l_1$ ($l_2$) represents the $l_1$($l_2$) on-average argument stability and $C_1-C_6$ are  positive constants. 
$m_1$ is the number of validation samples; 
$K$ and $T$ are the total numbers of outer and inner iterations. 
Assume that the output model has a small empirical risk $\mathbb{E}\left[R_{D_{m_1}}(A(D_{m_1},D_{m_2}))\right]= \mathcal{O}(m_1^{-1})$.
}
\scalebox{0.94}{ 
\begin{tabular}{c|c|ccccccccc}
\hline
\centering
Algorithms &\centering Stability &\centering SC-SC &\centering C-C &\centering NC-NC   \tabularnewline\hline
SSGD  
& $l_1$ 
& $\mathcal{O}\left( \frac{K}{m_1}\right)$ &  $\mathcal{O}\left( \frac{K^{C_4} \ln(K)}{m_1}\right)$ & ---  \\
(Theorem \ref{theorem2})& $l_2$ 
& $\mathcal{O}\left( \frac{(m_1+K)K}{m_1^2} \right)$ &  $\mathcal{O}\left( \frac{(m_1+K) K^{2C_4-1}\ln^2(K)}{m_1^2}\right)$ &---  \\
\midrule
\centering TSGD
& $l_1$ 
& $\mathcal{O}\left( \frac{K T^{C_5}}{m_1}\right)$ &  $ \mathcal{O}\left( \frac{\sqrt{2}^K T^{C_6} \ln(T)}{m_1 K}\right)$  & $\mathcal{O}\left( \frac{\sqrt{2}^K K^{C_2 T^{1+C_3}}T}{m_1}\right)$  \\
(Theorem \ref{theorem3})& $l_2$ 
& $\mathcal{O}\left( \frac{ (m_1+K) K T^{2C_5}}{m_1^2} \right)$ &  $\mathcal{O}\left( \frac{(m_1+K) 2^K  T^{2C_6} \ln^2(T)}{m_1^2 K^2}\right)$ & $\mathcal{O}\left( \frac{(m_1+K) 2^K K^{2C_2T^{1+C_3}}T^2}{m_1^2 }\right)$  \\
\midrule
SSGD 
& $l_1$ 
& $\mathcal{O}\left( \frac{1}{m_1}\right)$ &  $\mathcal{O}\left( \frac{1}{m_1}\right)$ & $\mathcal{O}\left( \frac{K^{C_1} }{m_1}\right)$  \\
(Proposition 1)& $l_2$ 
& $\mathcal{O}\left( \frac{m_1+K}{m_1^2 \sqrt{K}} \right)$ &  $\mathcal{O}\left( \frac{m_1+K}{m_1^2 \sqrt{K}}\right)$ &$\mathcal{O}\left( \frac{(m_1+K)K^{2C_1}}{m_1^2} \right)$  \\
\midrule
TSGD  & $l_1$ 
& $\mathcal{O}\left( \frac{K }{m_1}\right)$ & $\mathcal{O}\left( \frac{\sqrt{2}^K}{m_1 K}\right)$  & $\mathcal{O}\left( \frac{\sqrt{2}^K K^{C_2 T^{C_3}} T}{m_1}\right)$  \\
(Proposition 2)& $l_2$ 
& $\mathcal{O}\left( \frac{ (m_1+K) K )}{m_1^2} \right)$ &  $\mathcal{O}\left( \frac{(m_1+K) 2^K }{m_1^2 K^2}\right)$ & $\mathcal{O}\left( \frac{(m_1+K)2^K K^{C_2 T^{C_3}} T^2}{m_1^2}\right)$ 
\tabularnewline 
\hline
\end{tabular}}
\label{t1_highlight}
\end{table*}

Now we state the upper bounds of on-average argument stability for SSGD in Algorithm \ref{algorithm-SSGD}. 

\begin{theorem} \label{theorem2}
Suppose that Assumptions \ref{assumption_1}, \ref{assumption_2} hold and Algorithm $A$ is SSGD with $K$ iterations. Denote $\ell=\max \left\{\ell_f, \ell_g\right\}$, $\eta = \max\{ \eta_x, \eta_y\}$. 

(\textrm{I}) Assume that the bilevel optimization problem \eqref{bilevel_problem} is SC-SC with strong convexity parameters $\mu_f$ and $\mu_g$. Let the step sizes satisfy that $\frac{2(\mu_f+\mu_g)-\sqrt{4(\mu_f+\mu_g)^2-2\left(\ell_f^2+\ell_g^2\right)}}{2\left(\ell_f^2+\ell_g^2\right)} \leq \eta_x=\eta_y \leq \frac{2(\mu_f+\mu_g)+\sqrt{4(\mu_f+\mu_g)^2-2\left(\ell_f^2+\ell_g^2\right)}}{2\left(\ell_f^2+\ell_g^2\right)}$. 
Then, $A$ is $l_1(\beta)$ on-average argument-stable in expectation with
\begin{equation*}\small
\begin{aligned}
\beta=\frac{2C}{m_1} \sum_{k=1}^K \sqrt{2  \ell_f  E_{A, D_{m_1}} [R_{D_{m_1}}\left(x_k,y_k\right)]+L_g^2}
\end{aligned}
\end{equation*}
and 
$l_2(\beta^2)$ on-average argument-stable in expectation with
\begin{equation*}\small
\begin{aligned}
\beta^2= \frac{4(m_1+K) e C^2}{m_1^2} \sum_{k=1}^K (2  \ell_f E_{A, D_{m_1}} [R_{D_{m_1}}\left(x_k,y_k\right)]+ L_g^2),
\end{aligned}
\end{equation*}
where $C=\frac{2(\mu_f+\mu_g)+\sqrt{4(\mu_f+\mu_g)^2-2\left(\ell_f^2+\ell_g^2\right)}}{2\left(\ell_f^2+\ell_g^2\right)}$.

(\textrm{II}) Assume that the bilevel optimization problem \eqref{bilevel_problem} is C-C. If 
$\eta \leq \frac{c_1 \ln(K)}{\sqrt{2}K\ell}$  for some  $c_1 >0$, then $A$ is $l_1(\beta)$ on-average argument-stable in expectation with $\beta=$
\begin{equation*}\small
\begin{aligned}
\frac{\sqrt{2} c_1 \ln(K) K^{c_1 -1 }}{m_1  \ell} \sum_{k=1}^K \sqrt{2  \ell_f  E_{A, D_{m_1}} [R_{D_{m_1}}\left(x_k,y_k\right)]+L_g^2}.
\end{aligned}
\end{equation*}

And $A$ is $l_2(\beta^2)$ on-average argument-stable in expectation, where $\beta^2=$
\begin{equation*}\scriptsize
\begin{aligned}
\frac{2c_1^2(m_1+K) e K^{2c_1-2} \ln^2(K)}{ m_1^2 \ell^2} \sum_{k=1}^K (2  \ell_f E_{A, D_{m_1}} [R_{D_{m_1}}\left(x_k,y_k\right)]+ L_g^2).
\end{aligned}
\end{equation*}
\end{theorem}

\begin{remark}
Theorem \ref{theorem2} demonstrates that the algorithmic stability can be improved when the model can achieve a relatively small optimization error.
In addition, the $\ell_f$-smooth assumption can also be replaced by a H$\ddot{o}$lder continuous condition.
To obtain tighter bounds for the SSGD algorithm, we further analyze its algorithmic stability with refined step sizes in Proposition 1 of Appendix C.
\end{remark}

Combining Theorems \ref{theorem1} and \ref{theorem2}, the algorithmic generalization bounds of SSGD are further summarized in Table \ref{t1_highlight} under the low noise settings (small empirical risk). As shown in Table \ref{t1_highlight}, the generalization bounds of some SSGD algorithms achieve the rate of $\mathcal{O}(m_1^{-1})$ under the limitations of step sizes in Theorem \ref{theorem2}.  
From Table \ref{t1_highlight}, one can easily find that objective functions with better (convexity) properties usually lead to better algorithmic stability and generalization performance, which is consistent with the existing stability and generalization analysis for single-level problems \citep{DBLP:conf/icml/LeiY20,kuzborskij2018data,DBLP:conf/icml/LeiYYY21}.

\subsection{Stability and Generalization Analysis for TSGD}
Now we turn to establish the stability bounds of the TSGD algorithm with different inner and outer functions (i.e., NC-NC, C-C and SC-SC). 

Assume that $f$ is $\ell_f$-smooth and $g$ is $\ell_g$-smooth. Let $\eta_x$ and $\eta_y$ be the step sizes for updating $x$ and $y$, respectively. Denote $\nabla_y g(x,y)$ as the partial derivative of the function $g$ over the variable $y$. $y_K^t$ represents the inner parameter $y$ in $K$-th outer loop and $t$-th inner loop. For the TSGD algorithm $A$ with $K$ outer iterations and $T$ inner iterations, the argument stability $\left\|A(D_{m_1},D_{m_2})-A\left(D_{m_1}^{(i)},D_{m_2}\right)\right\|_2$ is measured by $\sqrt{\|x_{K}-x_{K}^{(i)}\|_2^2 +  \|y_{K}^0-(y_{K}^{0})^{(i)}\|_2^2}$, where
\begin{equation*}\small 
y_K^0 = y_{K-1}^T = y_{K-1}^0 - \sum_{t=0}^{T-1} \eta_y \nabla_y g(x_{K-1},y_{K-1}^t).
\end{equation*} 
\begin{remark}
Analogous to TSGD algorithm, the UD algorithm employed in \citep{DBLP:conf/nips/BaoWLZZ21} (see Algorithm \ref{algorithm-UD}) also involves two layers of nested loops but requires re-initialization in the inner level before each new outer loop. In their stability analysis, the inner-level parameter updates are not considered, but used to determine the constants of Lipschitz continuity and smoothness of the outer-level function.
This paper considers the general TSGD algorithms where both inner-level and outer-level parameters are updated continuously, e.g. $y_{K-1}^T = y_K^0$. The gradient summation of the inner-level parameter is relatively complex, making it difficult to directly utilize the (smooth or convex) properties \citep{DBLP:conf/nips/BaoWLZZ21,DBLP:conf/icml/HardtRS16}, which poses challenges for stability analysis.
\end{remark}

\begin{theorem} \label{theorem3}
Suppose that Assumptions \ref{assumption_1} and \ref{assumption_2} hold and algorithm $A$ is TSGD with $T$ inner loops and $K$ outer loops. Denote $\ell=\max\{\ell_f,\ell_g\}$, $\eta = \max\{ \eta_x ,\eta_y\}$, $E[R_{D_{m_1}}]=E_{A, D_{m_1}} [R_{D_{m_1}}\left(x_k,y_k\right)]$.

\textbf{(I)} Assume that the bilevel optimization problem is SC-SC with strong convexity parameters $\mu_f$ and $\mu_g$. 

Let the step sizes satisfy that 
$\frac{2(T\mu_g + \mu_f -T \ell)-\sqrt{4(T\ell-\mu_f-T\mu_g)^2 - 2(1+T^2)\ell^2(1-\frac{c_1\ln(T)}{K})}}{2(1+T^2)\ell^2} \leq \eta \leq \frac{2(T\mu_g + \mu_f -T \ell)+\sqrt{4(T\ell-\mu_f-T\mu_g)^2 - 2(1+T^2)\ell^2(1-\frac{c_1 \ln(T)}{K})}}{2(1+T^2)\ell^2}=C_1$ for some positive constant $c_1, C_1$.
Then $A$ is $l_1(\beta)$ on-average argument-stable in expectation with
\begin{equation*}\small
\begin{aligned}
\beta= \frac{2  T^{\frac{c_1}{2}} C_1}{m_1} \sum_{k=1}^K \sqrt{2  \ell_f  E[R_{D_{m_1}}]+T^2 L_g^2}
\end{aligned}
\end{equation*}
and $l_2(\beta^2)$ on-average argument-stable with 
\begin{equation*}\small
\begin{aligned}
\beta^2=\frac{4(m_1+K) e T^{c_1} C_1^2}{m_1^2} \sum_{k=1}^K (2  \ell_f E[R_{D_{m_1}}]+ T^2 L_g^2).
\end{aligned}
\end{equation*}

(\textrm{II}) Assume that the bilevel optimization problem is C-C. 
When $\eta \leq \frac{c_2 \ln(T)}{\sqrt{1+T^2} K \ell}$ for some $c_2>0$, $A$ is $l_1(\beta)$ on-average argument-stable in expectation with 
\begin{equation*}\small
\begin{aligned}
\beta=\frac{2 c_2 \ln(T) T^{c_2}}{m_1 \sqrt{1+T^2}K \ell} \sum_{k=1}^K 
2^{\frac{K-k}{2}}\sqrt{2  \ell_f  E[R_{D_{m_1}}]+T^2 L_g^2},
\end{aligned}
\end{equation*}
and is $l_2(\beta^2)$ on-average argument-stable with $\beta^2=$
\begin{equation*} \small
\begin{aligned}
\frac{4 c_2^2 (K+m_1) \ln^2(T) T^{2c_2} e  }{m_1^2 (1+T^2) K^2 \ell^2} \sum_{k=1}^K 2^{K-k} (2  \ell_f E[R_{D_{m_1}}]+ T^2 L_g^2).
\end{aligned}
\end{equation*}

(\textrm{III}) Assume that the bilevel optimization problem is NC-NC. Denote $\eta_{y,t}$ as the inner step size in $t$-th inner loop, denote $\eta_k=\max\{\eta_{x,k},\eta_{y,k}\}$ as the outer step size in $k$-th outer loop. Let $\eta_{y,t}\leq \frac{c_3}{\ell_g(t+1)}$, $\eta_k \leq \frac{c_4}{\ell k}$ and $\sum_{x=a}^b f(x) \leq c_5 \int_a^b f(x) d x$ for some positive constants $c_3,c_4, c_5$, then $A$ is $l_1(\beta)$ on-average argument-stable in expectation with
\begin{equation*}\small
\begin{aligned}
\beta=\frac{2 c_4 }{ m_1 \ell}\sum_{k=1}^K 2^{\frac{K-k}{2}} (\frac{K}{k})^{c_4 c_5 T^{c_6}\sqrt{1+T^2}} \sqrt{ 2  \ell_f E[R_{D_{m_1}}] + T^2 L_g^2},
\end{aligned}
\end{equation*}
and $l_2(\beta^2)$ on-average argument-stable with $\beta^2= $
\begin{equation*}\scriptsize
\begin{aligned}
&\frac{4(m_1+K) c_4^2 e}{ m_1^2 \ell^2}\sum_{k=1}^K  (\frac{K}{k})^{2 c_4 c_5 T^{c_6}\sqrt{1+T^2}} 2^{K-k}  (2  \ell_f E[R_{D_{m_1}}]+ T^2 L_g^2 ).
\end{aligned}
\end{equation*}

Notice that $T^{c_6}\sqrt{1+T^2}$ with $\eta_{y,t} \leq \frac{c_3}{\ell_g(t+1)}$ is obtained from Lemma 6 for NC-NC in \emph{Appendix D}, where the original form is $T^{c_0 c_1}\sqrt{1+T^2}$ with $\eta_{y,t} \leq \frac{c_0}{\ell_g(t+1)}$. 

\end{theorem} 

\begin{remark} After integrating Theorems \ref{theorem1} and \ref{theorem3}, we summarize the generalization bounds of Algorithm \ref{algorithm-TSGD} in Table \ref{t1_highlight}.  Similar to Theorem \ref{theorem2}, the results of Theorem \ref{theorem3} also demonstrate that the total numbers of the validation samples $m_1$ ($\uparrow$), the inner iterations $T$ ($\downarrow$) and outer iterations $K$ ($\downarrow$) directly affect the generalization performance ($\uparrow$) of TSGD algorithms. We also observe that the impacts of $K$ and $T$ on generalization are suppressed for Algorithm \ref{algorithm-TSGD} with SC-SC (or C-C) with a small enough step size. 
In order to obtain tighter bounds w.r.t. Theorems \ref{theorem2} and \ref{theorem3}, we further derive the corresponding results with refined step sizes in Propositions 1 and 2 in Appendix C, D.
The results shown in Table \ref{t1_highlight} are comparable to \cite{DBLP:conf/nips/BaoWLZZ21} with the bound of $\mathcal{O}(\frac{K^c}{m})$ where $0<c<1$. Relaxing the stepsize limitations, especially for SC-SC, is a meaningful direction for future work.
\end{remark}


\begin{figure*}[!t]
\centering
\subfigure[Validation Error]{\includegraphics[width = 0.28\textwidth]{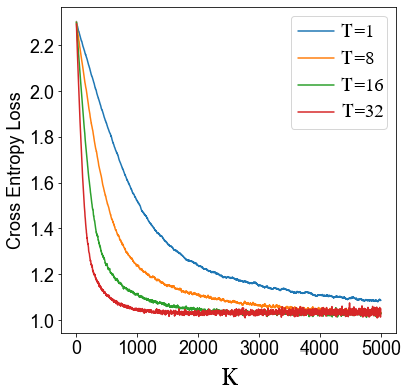}\label{f1_1}}
\subfigure[Testing Error]{\includegraphics[width = 0.28\textwidth]{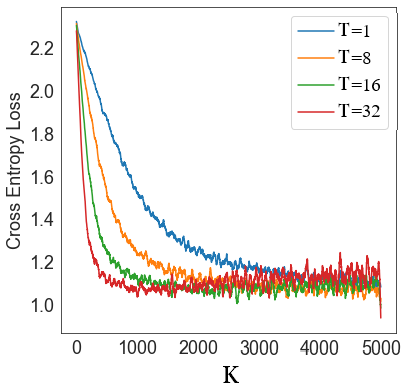}\label{f1_2}}
\subfigure[Generalization Gap]{\includegraphics[width = 0.287\textwidth]{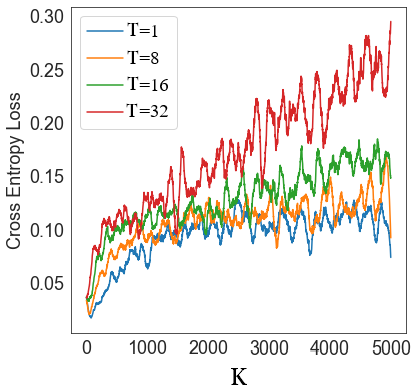}\label{f1_3}}
\caption{Results of hyperparameter optimization in data reweighting with varying $T$ and $K$}
\label{figure_1}
\end{figure*}
\begin{figure*}[!t]
\centering
\subfigure[Validation Error]{\includegraphics[width = 0.28\textwidth]{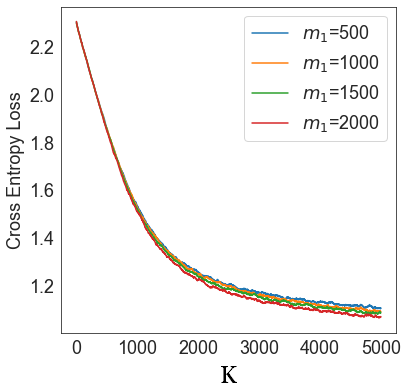}\label{f2_1}}
\subfigure[Testing Error]{\includegraphics[width = 0.28\textwidth]{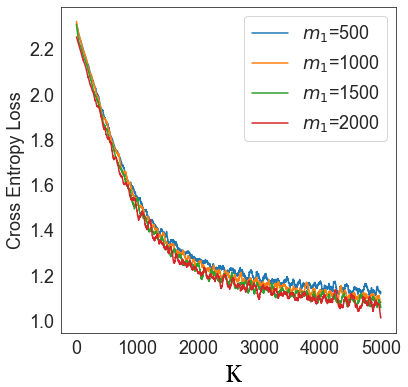}\label{f2_2}}
\subfigure[Generalization Gap]{\includegraphics[width = 0.287\textwidth]{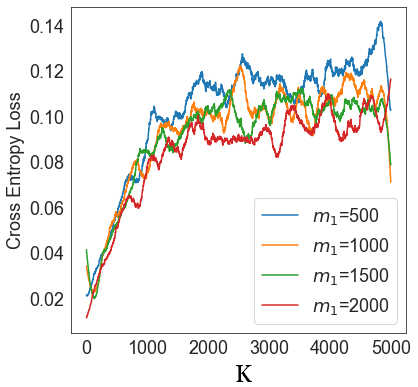}\label{f2_3}}
\caption{Results of hyperparameter optimization in data reweighting with varying $K$ and $m_1$}
\label{figure_2}
\end{figure*}

\section{Empirical Evaluations} \label{section5}
This section empirically validates our theoretical findings on two real-world datasets. We consider Algorithm \ref{algorithm-TSGD} here, since it is equal to Algorithm \ref{algorithm-SSGD} as $T=1$. The distributions of the test and validation samples are assumed to be the same, but may differ from those of the training data \citep{ren2018learning,DBLP:conf/nips/BaoWLZZ21}. Similar to \citep{DBLP:conf/nips/BaoWLZZ21}, we focus on evaluating the generalization behavior of outer-level problems based on the validation set. 
All experiments are implemented in Python on an Intel Core i7 with 32 GB of memory. Implemented codes (including \citep{DBLP:conf/nips/BaoWLZZ21} for hyperparameter optimization) and data sets (including the MNIST data \citep{lecun1998mnist} and the Omnilot data \citep{science15_omnilot})  are from publicly available sources.

This section considers the general hyperparameter optimization formulation \citep{hyper_convergence1}. Given the training set $D_{\text {train}}$ and the validation set $D_{\text {val}}$, the hyperparameter optimization scheme can be  formulated as
\begin{equation*}\small
\begin{aligned}
&\min_x \mathcal{R}_{D_{\text {val }}}(x)=\frac{1}{\lvert D_{\text {val }}\lvert} 
\sum_{\xi \in D_{val}} h\left(x,y^*(x); \xi\right) \\
\text {~s.~t.~ } &  y^*=\underset{y}{\arg \min } \underbrace{\frac{1}{\lvert D_{\text {train }}\lvert} \sum_{\zeta \in D_{\text {train }}}\left(h(x,y;\xi)+\Omega_{y, x}\right)}_{\mathcal{R}_{D_{\text {train }}}(x, y)},
\end{aligned}
\end{equation*}
where $h$ is the loss function, ${\Omega}_{y, x}$ is the regularizer and $\lvert D_{\text {train }}\lvert$ represents the size of training data. 

\subsection{Experiment Settings}
We evaluate the impact of several factors on the generalization gap \eqref{generalization_gap} using the famous MNIST dataset \citep{lecun1998mnist}, which consists of more than $6\times 10^5$ handwritten digits with a size of $28 \times 28$. 
Following the same data reweighting task in \citep{DBLP:conf/nips/BaoWLZZ21}, we randomly corrupt the labels of training samples with a probability of $50\%$ and employ a fully connected network (with sizes 784/256/10) with cross-entropy loss for classification. Initially, we randomly select 2000, 2000, and 1000 figures for training, validation and testing, respectively. Meanwhile, set the initial batch size to 8, the maximum number of inner iterations to $T=5000$, and the number of outer iterations to $K=5000$. The initial step sizes for inner and outer minimization problems are 0.01 and 5, respectively. For the given parameter settings, each experiment is repeated 5 times on a single GeForce GTX 1660 SUPER GPU, and the average results are reported.

\subsection{Experimental Results}
The generalization gap defined in \eqref{generalization_gap} is estimated by the divergence between the validation error and the testing error. 

\emph{Impact of iteration numbers $K$ and $T$.} Now we evaluate the impact of parameters (e.g., the number of validation samples $m_1$, inner iteration $T$, and outer iteration $K$) on the generalization performance.
Figure \ref{figure_1} shows the curves of validation error, testing error, and the generalization gap under different settings of maximum inner iteration $T$ and maximum outer loop $K$.
Figures \ref{f1_1} and \ref{f1_2} imply that the classification model might be overfitting with increasing testing errors as $K>3000$ and $T=32$. 
Besides, Figure \ref{f1_3} shows that overly large $K$ and $T$ can reduce the generalization ability of the hyperparameter optimization method due to overfitting. This empirical finding is consistent with our theoretical results and the previous related analysis \citep{franceschi2018bilevel,DBLP:conf/nips/BaoWLZZ21}. 

\emph{Impact of sample size $m_1$ with $T=1$.}
Figure \ref{figure_2} presents the results of SSGD (Algorithm \ref{algorithm-SSGD}) under different choices of $K$ and $m_1$. 
From Figure \ref{f2_3}, we observe that a small sample size $m_1=500$ leads to an increase in validation error and testing error. This indicates that a larger number of validation samples is beneficial for reducing the generalization gap. 
The above empirical findings match our theoretical results, see e.g., Theorem \ref{theorem3} and Table \ref{t1_highlight}.

Based on theoretical analysis and empirical evaluations, we can get some understanding of the generalization performance of bilevel optimization.  Explicitly, the generalization ability of SBO can often be improved by increasing $m_1$ and the proper iteration numbers $K$ and $T$, where too small iterations may cause underfitting and too large ones can lead to overfitting. Usually, it is beneficial for generalization to set appropriate learning rates, especially in NC-NC. In real applications, the trade-off among $m_1$, $T$, and $K$ is crucial for ensuring the effectiveness of SBO methods.

\section{Conclusion} \label{section6}
This paper establishes stability and generalization analyses for stochastic bilevel optimization using first-order gradient-based approximate algorithms. 
Our theoretical results are obtained by developing an analysis technique for the on-average argument stability and can cover a wider range of bilevel optimization algorithms under low-noise settings. 
Compared with the state-of-the-art analysis \citep{DBLP:conf/nips/BaoWLZZ21}, our theoretical results do not require reinitializing the inner-level parameter before each iteration and are suitable for objective functions under milder conditions. 

\section*{Acknowledgments}
This work was supported in part by the National Natural Science Foundation of China (Nos. 62376104 and 12071166), the Fundamental Research Funds for the Central Universities of China (No. 2662023LXPY005), and HZAU-AGIS Cooperation Fund (No. SZYJY2023010).




\bibliographystyle{named.bst}
\bibliography{ijcai24}

\end{document}